%% file: ms.tex
\documentclass[11pt,a4paper]{article}
\usepackage[dvipsnames]{xcolor}

\usepackage[hyperref]{acl2018}
\usepackage{times}
\usepackage{latexsym}

\usepackage{url}
\usepackage{amsmath}
\usepackage{amssymb}
\usepackage{xfrac}
\usepackage{bm}
\usepackage{physics}
\usepackage{scalerel}
\usepackage{booktabs,siunitx}
\usepackage{rotating}
\usepackage{tikz}
\usepackage{setspace}
\usetikzlibrary{shapes,arrows,positioning,bayesnet,shadows}
\usepackage{verbatim}
\usepackage{subcaption}
\usepackage{url}
\usepackage[utf8]{inputenc}

\input{macros}

\aclfinalcopy

\title{Modeling Latent Sentence Structure in Neural Machine Translation}

\author{Jasmijn Bastings\affmark[1] \quad Wilker Aziz\affmark[1] \quad Ivan Titov\affmark[1,2] \quad \textbf{Khalil Sima'an\affmark[1]} \\
       \affaddr{\affmark[1]ILLC, University of Amsterdam} \quad
\affaddr{\affmark[2]ILCC, University of Edinburgh}\\
        {\tt \{bastings,titov,w.aziz,marcheggiani,k.simaan\}@uva.nl}
}

\date{}

\begin{document}
\maketitle
\input{abstract}
\input{introduction}
\input{model}

\input{experiments}

\input{discussion}

\input{related}
\input{conclusion}

\section*{Acknowledgments}
This work was supported by the European Research Council (ERC StG BroadSem 678254) and the Dutch National Science Foundation (NWO VIDI 639.022.518, NWO VICI 277-89-002).

\bibliography{acl}
\bibliographystyle{acl_natbib}

\end{document}

%% file: macros.tex
\newcommand*{\affmark}[1][*]{\textsuperscript{#1}}
\newcommand*{\affaddr}[1]{#1} 

\newcommand{\dimension}[2]{{\ensuremath{#1_{\textsc{#2}}}}}
\newcommand{\mparam}[2]{\ensuremath{\mathbf{#1}_{\!\textsc{#2}}\,}}

\DeclareMathOperator{\Cat}{Cat}

\DeclareMathOperator{\Concrete}{Concrete}

%% file: abstract.tex
\begin{abstract}

Recently it was shown that linguistic structure predicted by a supervised parser can be beneficial for neural machine translation (NMT). In this work we investigate a more challenging setup: we incorporate sentence structure as a latent variable in a standard NMT encoder-decoder and induce it in such a way as to benefit the translation task. 
We consider German-English and Japanese-English translation benchmarks and observe that when using RNN encoders the model makes no or very limited use of the structure induction apparatus. In contrast, CNN and word-embedding-based encoders rely on latent graphs and force them to encode useful, potentially long-distance, dependencies.\footnote{Accepted as an extended abstract to ACL NMT 2018}
\end{abstract}

%% file: introduction.tex
\section{Introduction}

Recently it was shown that syntactic structure can be beneficial for neural machine translation (NMT)~\citep{eriguchi2016treetoseq,hashimoto-tsuruoka:2017:EMNLP2017,BastingsEtAl2017GCN}. For example, \citet{BastingsEtAl2017GCN} used graph convolutional networks to encode  linguistic inductive bias about syntactic structure of the source sentence. Instead of relying on supervised parsers, in this work we consider a more challenging setting: we incorporate sentence structure as a latent variable in a standard NMT encoder-decoder and induce it in such a way as to benefit the translation task. 

Inducing latent structure while incurring a downstream loss was explored for e.g. sentiment analysis and textual entailment \citep{yogatama2017iclr,maillard2017jointlylearning,choi2017unsupervisedtree,kim2017structuredattention}.
Interestingly, \citet{williamsbowman2017isitsyntax} showed that these learned structures do not correspond to syntactic/semantic generalizations, but can be as useful as access to predicted parses.

Our goal is to investigate under which conditions induced latent structures can be beneficial for NMT.
Although we would like these structures to be discrete (e.g. for better interpretability), we do not enforce discreteness in order to avoid high-variance estimators. Instead, we induce structure in the form of weighted densely-connected graphs.

We design our probabilistic model with two components (see Figure~\ref{fig:architecture}): (1) a \textit{graph component} that stochastically samples a latent graph conditioned on the source sentence, and (2) a graph-informed \textit{translation component} that conditions on the sampled graph and the source sentence to predict the target sentence using a recurrent decoder. 
The graph component is modeled as a Concrete distribution \citep{MaddisonEtAl2017:Concrete,JangEtAl2017:GumbelSoftmax}, thus promoting graphs that are approximately discrete. The graph-informed component uses graph convolutional networks, in a similar way as in \citet{BastingsEtAl2017GCN}, but relying on latent graphs instead of syntactic parsers.

Using two distinct components lets us disentangle their effects and study in which conditions 
useful structure gets induced. 
To that end, we keep the architecture of the graph component fixed across experiments and vary the encoder of the translation component (e.g. RNN, CNN, or embeddings). We observe that with RNNs, likely due to their expressiveness, the model makes no or very limited use of the latent graph apparatus. In contrast, with CNN  encoders the model finds purpose to latent graphs such as encoding useful, potentially long-distance, dependencies in the source sentences.

Our contributions are threefold: (1) we formulate an architecture with two components that stochastically induces approximately discrete source-side graphs; (2) we study how 
varying the encoder type influences the resulting latent graphs; (3) we validate our approach on En-De and En-Ja.

%% file: model.tex
\section{\label{sec:model}Model}

\input{fig_architecture}

Our model is a deep generative model; that is, a probabilistic model whose components are parameterized by neural networks. 
There are two such probabilistic components: (1) a graph sampler and (2) a translation component. Both components require some form of encoder, while sharing word embedding matrices.

\subsection{Graph Component\label{sec:graph-component}}

The graph component conditions on the source sentence $x_1^m$ and samples for each source position $i$ an $m$-dimensional probability vector
\begin{equation}
A_i|x_1^m \sim \Concrete(\tau, \boldsymbol\lambda_i)
\end{equation}
whose $k$th component $a_{ik}$ represents the relative strength of the edge from $x_i$ to $x_k$.
Then, altogether, $a_1^m$ can be seen as the adjacency matrix of a weighted fully-connected graph over the source words.
By analogy to dependency parsing, we can see each $a_i$ as the parameter vector of a Categorical distribution over the candidate \emph{heads} of $x_i$, which is why we call the Concrete parameter $\boldsymbol\lambda_i \in \mathbb R^m$ a vector of \emph{head potentials}. 
Given a sequence of source word embeddings, we obtain hidden states $\hat{\mathbf s}_1^m$ using a bi-directional LSTM \citep{hochreiter97,schusterpaliwal1997}. From these hidden states, we then create `key' and `query' (or `head' and `dependent', by analogy) representations for each state $\hat{\mathbf s}_i$ using linear projections:
\begin{equation}
\begin{aligned}
\mathbf k_i = \mparam{W}{k} \hat{\mathbf s}_i \qquad
\mathbf q_i = \mparam{W}{q} \hat{\mathbf s}_i
\end{aligned}
\end{equation}
with $\mparam{W}{k}, \mparam{W}{q} \in \mathbb{R}^{\dimension{d}{k} \times d}$. 

\noindent We then obtain head potentials using a scaled dot product:
\begin{equation}
\begin{aligned}
\lambda_{ik} = 
  \begin{cases} 
   \frac{1}{\sqrt[]{\dimension{d}{k}}} \mathbf q_i^\top \mathbf k_k & \text{if } i \neq k \\
   -\infty       & \text{if } i=k
  \end{cases}
\end{aligned}
\end{equation}
Similar projections are used by \citet{dozat2017biaffine} and \citet{vaswani2017attention}. Importantly, they break the symmetry of the dot product, which is crucial to model a \emph{directed} graph.\footnote{We mask out the diagonal ($i=k$) to demote induction of trivial edges (from a word to itself).} 

The Concrete density also takes a temperature parameter $\tau$ which we made a global parameter. We describe our decaying scheme in \S\ref{sec:experiments}.

\subsection{Translation Component\label{sec:translation-component}}

The translation component conditions on the source sentence $x_1^m$, a sampled graph $a_1^m$, and a target prefix $y_{<j}$ to sample a target word
\begin{equation}
Y_j|x_1^m, a_1^m, y_{<j} \sim \Cat(\boldsymbol\pi_j) 
\end{equation}
at each time step $j$. 

To do so, we have an attention-based encoder-decoder similar to that of \citet{BastingsEtAl2017GCN} compute the Categorical parameters $\boldsymbol\pi_j$ at each time step. 
We first obtain an encoding $\mathbf s_1^m$ of the source sentence, which is independent of the representations used by the graph component, 
and then use graph convolutions to enhance these representations given the neighborhood defined by the graph $a_1^m$.
After obtaining such enriched representations we employ a standard attentive decoder. 

\paragraph{Encoder.} 
We experiment with three different encoders for the translation component. 
In the simplest case we use word embeddings and add \textit{position encodings} to them; we use time series as proposed by \citet{vaswani2017attention}.
We also use convolutional layers as also used by \citet{gehring-EtAl:2017:Long}, and again add position encodings.
Lastly, we use a bi-directional RNN as used in \citet{bahdanau15iclr}. We use LSTMs as our RNN cells.

\paragraph{Graph Convolution.} 
We now employ the graph convolutional networks of \citet{marcheggiani-titov:2017:EMNLP2017} and \citet{BastingsEtAl2017GCN} to incorporate graph $a_1^m$ into source word representations $\mathbf s_1^m$:
\begin{equation}
\mathbf s_i = \text{GCN}(\mathbf s_1^m, a_1^m)[i]
\end{equation}
Since we induce unlabeled graphs, we do not use any label-specific GCN parameters.
Note that the GCN creates an elegant interface between the graph component and the translation component which prevents the former from ``leaking'' parameters or representations (except $a_1^m$) to the latter.

\paragraph{Decoder.}
Our decoder is based on \citet{luong-pham-manning:2015:EMNLP}; for the $j$th prediction an LSTM attends to the (graph-informed) source word representations.

\subsection{Parameter estimation}
We estimate the parameters of our model to maximize a lower bound on marginal likelihood \begin{equation}
\begin{aligned}\label{eq:objective}
	\int   p(a_1^m|x_1^m) \log p(y_1^n|x_1^m) \dd{a_1^m} 
\end{aligned}
\end{equation}
obtained by application of Jensen's inequality.
We get unbiased gradient estimates for this objective by sampling a single graph per source sentence. The Concrete density is a location family \citep{MaddisonEtAl2017:Concrete}, thus we can reparameterize samples from the graph component, which is essential to enable parameter estimation via backpropagation \citep{Kingma+2014:VAE}.

%% file: fig_architecture.tex
\begin{figure}[t]
\begin{tikzpicture}[node distance=1.3cm,auto,font={\sffamily\small}]
\tikzset{>=latex}
\tikzset{>=stealth}
\tikzstyle{graph-comp} = [draw=gray,fill=white,drop shadow={shadow xshift=1.5pt,shadow yshift=-1.5pt,color=lightgray}]
\tikzstyle{translation} = [draw=gray,fill=white,drop shadow={shadow xshift=1.5pt,shadow yshift=-1.5pt,color=lightgray}]

\tikzstyle{block} = [rectangle,fill=white,draw=gray,text width=8.2em, text centered, rounded corners, minimum height=2.0em,thick]
\tikzstyle{line} = [draw=gray,->,shorten >=2.2pt, shorten <=2.7pt,thick]

\draw [rounded corners,draw=Fuchsia,dashed] (-5.75,-0.5)--(-5.75,1.8)--(-2.13,1.8)--(-2.13,-0.5)--cycle;
\draw [rounded corners,draw=NavyBlue,dashed] (-1.8,-0.5)--(-1.8,3.1)--(1.82,3.1)--(1.82,-0.5)--cycle;

\node [block,translation] (encoder) {Encoder};
\node [block,above of=encoder,xshift=0cm,translation] (gcn) {Graph Convolution};
\node [block,above of=gcn,translation] (decoder) {Decoder};
\node [block,left=0.5cm of gcn,graph-comp] (graph) {Graph Sampler};
\node [block, below of=graph, graph-comp] (graph-encoder) {BiLSTM};
\path [line] (graph) -- (gcn);
\path [line] (encoder) -- (gcn);
\path [line] (gcn) -- (decoder);
\path [line] (graph-encoder) -- (graph);

\node[text=Fuchsia,node distance=0.9cm,below of=graph-encoder] {Graph component};
\node[text=MidnightBlue,node distance=0.9cm,below of=encoder] {Translation component};
\end{tikzpicture}
\caption{\label{fig:architecture}Schematic view of the architecture.}
\end{figure}
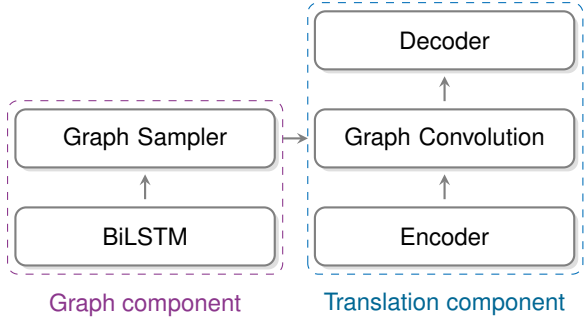

%% file: experiments.tex
\input{tab-data}

\input{table_results}

\section{Experiments}
\label{sec:experiments}

We build our models on top of Tensorflow NMT\footnote{\url{https://github.com/tensorflow/nmt}} and experiment on German$\leftrightarrow$English and Japanese$\leftrightarrow$English tasks. 
Data set statistics are summarized in Table~\ref{tab:data}.

\paragraph{De$\leftrightarrow$En.} We train on IWSLT14 with the same splits and pre-processing as \citet{ranzato2016sequence}.

\paragraph{Ja$\leftrightarrow$En.} We train on the Asian Scientific Paper Excerpt Corpus (ASPEC) \cite{nakazawa2016aspec} as pre-processed by the WAT 2017 Small-NMT task using SentencePiece.\footnote{\url{http://lotus.kuee.kyoto-u.ac.jp/WAT/WAT2017/snmt/index.html}}
We use the provided dev and test sets.

\subsection{Baselines}
For our baselines we train our models \emph{without} the graph sampler, varying the encoder. 
We add a dense layer with ReLU activation and residual connection on top of the encoder, to make our baselines stronger and to keep the number of parameters for the translation component equal.\footnote{This is identical to a GCN layer with self-loops only.}

\subsection{Hyperparameters}
We optimize using Adam \citep{kingma2015adam}.
For De-En, we use 256 hidden units, a learning rate of 3e-4, and dropout 0.3. For Ja-En, we use 512 units, a learning rate of 2e-4, and dropout 0.2. Word representations (query and key) are projected down to $d_k = 256$ units when calculating head potentials. Our batch size is set to 64. Beam search is used with beam size 10 and with a length penalty of 1.0. 

\paragraph{Concrete Temperature.}
For the graph component we define an initial temperature $\tau_0$ and apply exponential decay based on the number of network updates. After $t$ updates, the temperature is $\tau_0 \times d^{\lfloor t / t_d \rfloor}$ with decay rate $d$ and decay steps $t_d$.  
We set $\tau_0 = 2$, $d=0.99$, and $t_d$ 1 epoch.

\subsection{Evaluation}
We use Sacr\'eBLEU\footnote{\url{https://github.com/mjpost/sacreBLEU}} to report all BLEU scores.
For German-English we report case-sensitive tokenized BLEU scores to compare with previous work. For Japanese-English, we report detokenized BLEU for English using the 13a tokenizer (which is mteval-v13a compatible). For Japanese we report tokenized BLEU on the segmentation from SentencePiece in accordance with the Small-NMT shared task. 

\subsection{Results}
Table~\ref{tab:results} lists our results.
We observe that the baselines with LSTM encoders outperform the CNN ones, to be followed by the word embedding baselines.
This is not surprising, since the LSTM is the only baseline that can fully capture the context of a word. The CNN baseline, using position encodings, actually performs surprisingly well, despite having a receptive field of only five words.

We observe that substantial gains in BLEU score can be made when latent graphs are incorporated into models with word embedding and CNN encoders. 
This suggests that the graphs are capturing useful relations outside of the receptive fields of the encoders.
For the BiLSTM encoders the latent graphs do not seem beneficial overall. We look into this in the next section.

%% file: tab-data.tex
\begin{table}[t]
\centering
\begin{tabular}{lrrrr}
\toprule
     & \textsc{\small Train} & \textsc{\small Dev} & \textsc{\small Test} & \textsc{\small Vocabularies} \\
\midrule
De-En & 153K & 7282 & 6750 & 32010/22823 \\
Ja-En &   2M & 1790 & 1812 & 16384 (SPM)\\
\bottomrule
\end{tabular}
\caption{Data set statistics.}
\label{tab:data}
\end{table}

%% file: table_results.tex
\begin{table*}[tb]
\centering
\begin{tabular}{llrrrr}
\toprule
 & & \multicolumn{2}{c}{\textsc{\small IWSLT14}} & \multicolumn{2}{c}{\textsc{\small WAT17}} \\
 	\cmidrule(lr){3-4} 	\cmidrule(lr){5-6}
 & \textsc{\small Encoder} & \textsc{\small De-En} & \textsc{\small En-De} & \textsc{\small Ja-En} & \textsc{\small En-Ja} \\ \midrule
Ext. baseline   & RNN & 27.6     & -     & -    & 28.5\\
\midrule
Baseline 	& Emb. & 22.7 & 17.9 & 18.1 & 18.1 \\
Baseline 	& CNN  & 23.6 & 19.1 & 23.0 & 24.6 \\
Baseline 	& RNN  & 27.6 & 22.4 & 26.0 & 28.7 \\
\midrule
Latent Graph & Emb. & 24.0 & 18.7 & 23.2 & 24.3 \\
Latent Graph & CNN  & 24.6 & 20.3 & 24.6 & 26.7 \\
Latent Graph & RNN  & 27.2 & 22.4 & 26.0 & 29.1 \\
\bottomrule
\end{tabular}
\label{tab:results}
\caption{Test results for German$\leftrightarrow$English and Japanese$\leftrightarrow$English.}
\label{tab:results}
\end{table*}

%% file: discussion.tex
\input{table_head}

\section{Discussion}
What dependencies are the graphs capturing?
The analysis of our graphs is somewhat nontrivial as they capture dependencies over sub-word units, are not discrete, and lack gold-truth parse trees.

We first measure the distance between each word and its most-likely head word. 
If this distance is small on average, then words typically select their neighboring word as head, whereas if it is larger then this suggests potentially interesting non-local dependencies.
Table~\ref{tab:enja-head-distance} lists the average head distances for En-Ja, together with the variance over all distances.
We find that with LSTM-encoders words typically select their heads nearby, whereas with the other encoders heads are also found further away. Figure~\ref{fig:example-latent-graphs} indeed shows this for an example sentence. Inspection reveals that for the LSTM  the graphs became trivial, confirming that it already captures non-local dependencies. 

\input{table_entropy}

We also wonder how sparse our graphs are.
To find out, 
we interpret the adjacencies in the graph as Categorical head distributions and report average entropy (normalized by sentence length) in 
Table~\ref{tab:enja-entropy}.
If each word was to select its head uniformly, this would result in a value of $28.7$.
However, we observe much lower values, indicating that our graphs are in fact rather sparse.

\input{fig_example_graphs}

%% file: table_head.tex
\begin{table}[h]
\centering
\begin{tabular}{lcccc}
\toprule
 & \multicolumn{2}{c}{\textsc{\small Mean head distance}}\\
 	\cmidrule(lr){2-3}
\textsc{\small Encoder} & \textsc{\small Ja-En} & \textsc{\small En-Ja} &  \\ \midrule
Emb. & 4.0 \tiny$\pm 6.9$ & 3.8 \tiny$\pm 5.6$  \\
CNN  & 6.1 \tiny$\pm 6.5$ & 6.7 \tiny$\pm 7.1$  \\
RNN  & 4.3 \tiny$\pm 6.5$ & 2.0 \tiny$\pm 5.4$ \\
\bottomrule
\end{tabular}%
\caption{Mean head distance for En-Ja.}
\label{tab:enja-head-distance}
\end{table}

%% file: table_entropy.tex
\begin{table}[h]
\centering
\begin{tabular}{lcccc}
\toprule
 & \multicolumn{2}{c}{\textsc{\small Mean entropy}} \\
 	\cmidrule(lr){2-3}
\textsc{\small Encoder} & \textsc{\small Ja-En} & \textsc{\small En-Ja} \\ \midrule
Emb. & 0.49 \tiny$\pm 0.18$ & 0.42 \tiny$\pm 0.18$\\
CNN  & 1.21 \tiny$\pm 0.28$ & 1.47 \tiny$\pm 0.30$\\
RNN  & 0.51 \tiny$\pm 0.20$ & 0.00 \tiny$\pm 0.01$\\
\bottomrule
\end{tabular}%
\caption{Mean Entropy for En-Ja.}
\label{tab:enja-entropy}
\end{table}

%% file: fig_example_graphs.tex
\begin{figure*}[t]
    \begin{subfigure}[b]{0.33\textwidth}
        \includegraphics[width=4cm,trim={0 0 0 0},clip]{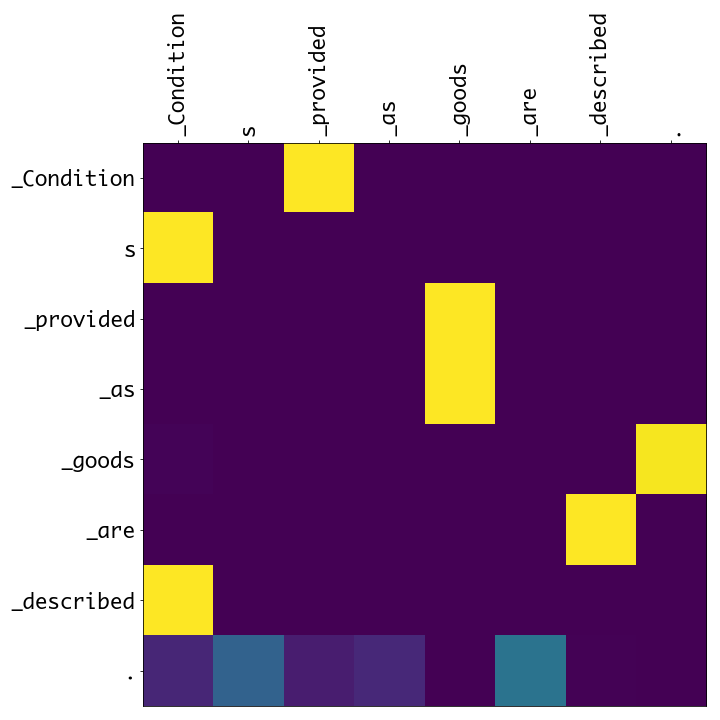}%
        \caption{En-Ja Emb}
    \end{subfigure}%
    \begin{subfigure}[b]{0.33\textwidth}
        \includegraphics[width=4cm,trim={0 0 0 0},clip]{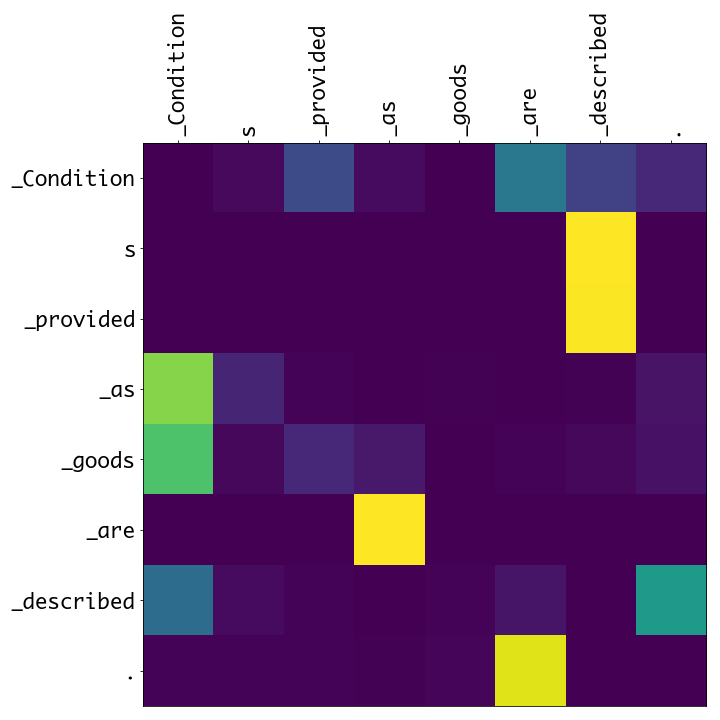}%
        \caption{En-Ja CNN}
    \end{subfigure}%
    \begin{subfigure}[b]{0.33\textwidth}
        \includegraphics[width=4cm,trim={0 0 0 0},clip]{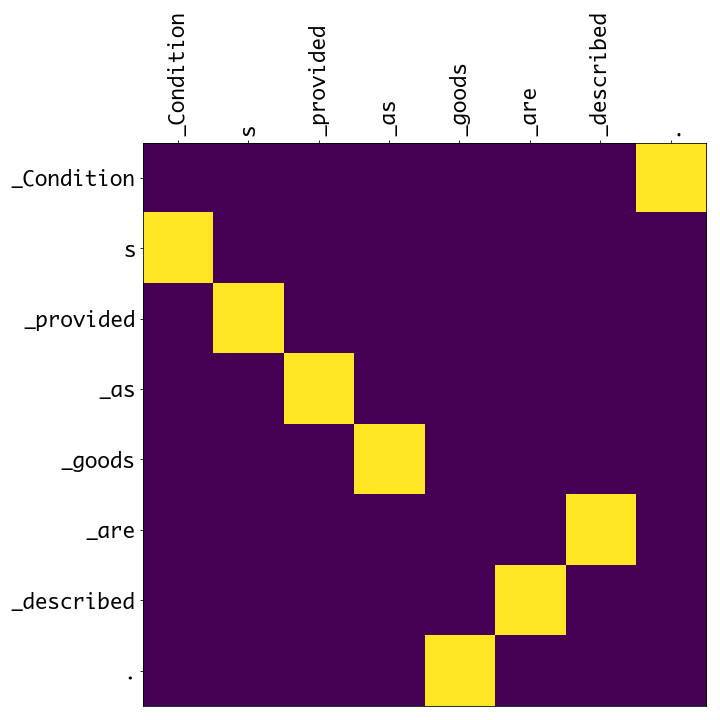}%
        \caption{En-Ja RNN}
    \end{subfigure}    \\
    \begin{subfigure}[b]{0.33\textwidth}
        \includegraphics[width=4cm,trim={0 0 0 0},clip]{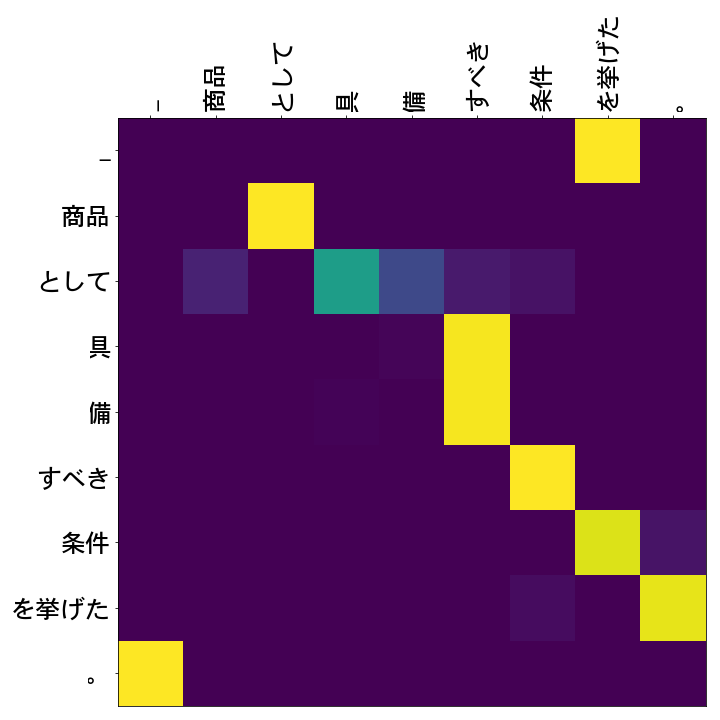}%
        \caption{Ja-En Emb}
    \end{subfigure}%
    \begin{subfigure}[b]{0.33\textwidth}
        \includegraphics[width=4cm,trim={0 0 0 0},clip]{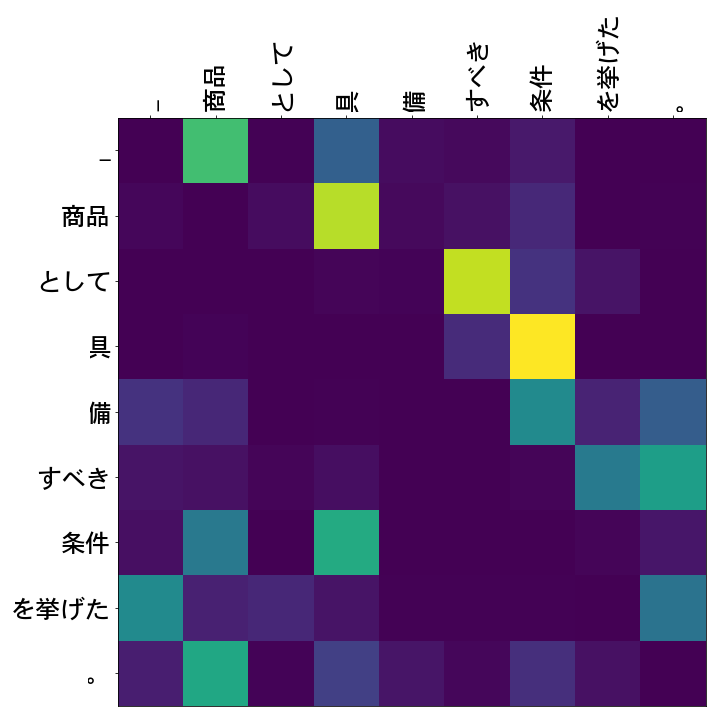}%
        \caption{Ja-En CNN}
    \end{subfigure}%
    \begin{subfigure}[b]{0.33\textwidth}
        \includegraphics[width=4cm,trim={0 0 0 0},clip]{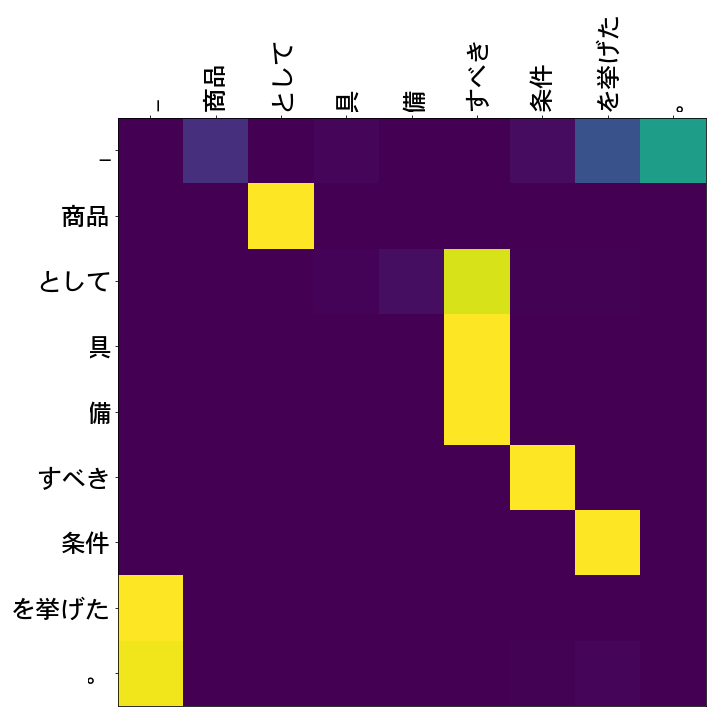}%
        \caption{Ja-En RNN}
    \end{subfigure}      
    \caption{Example latent graphs for En-Ja and Ja-En.}
    \label{fig:example-latent-graphs}
\end{figure*}

%% file: related.tex
\section{Related Work}
\citet{hashimoto-tsuruoka:2017:EMNLP2017} and \citet{tran2018inducing}
 induce relaxed graphs deterministically on the source side.
\citeauthor{hashimoto-tsuruoka:2017:EMNLP2017} use a vanilla self-attention mechanism, whereas \citeauthor{tran2018inducing} use structured attention.
Both do so on top of BiLSTM encodings and attend directly to a transformation of the same encodings and/or additional context vectors.
In this work, instead, we introduce a clear-cut separation that largely reduces the risk of over-parameterisation.
Our stochastic induction also opens the possibility to explore other sparsity induction priors (e.g. Dirichlet).
In contrast to e.g. \citeauthor{hashimoto-tsuruoka:2017:EMNLP2017}, we operate directly on sub-word sequences, eliminating word-level dependency pre-training.

%% file: conclusion.tex
\section{Conclusion}
We presented a model with separate graph induction and translation components and studied if our induced latent graphs are beneficial using three different encoders.
In the case of LSTM encoders the graphs turned out to be (largely) trivial, while for the simpler word embedding and CNN encoders they contain useful, potentially long-distance dependencies. 
\newpage

%% file: ms.bbl
\begin{thebibliography}{}
\expandafter\ifx\csname natexlab\endcsname\relax\def\natexlab#1{#1}\fi

\bibitem[{Bahdanau et~al.(2015)Bahdanau, Cho, and Bengio}]{bahdanau15iclr}
Dzmitry Bahdanau, Kyunghyun Cho, and Yoshua Bengio. 2015.
\newblock {Neural Machine Translation by Jointly Learning to Align and
  Translate}.
\newblock In {\em {Proceedings of the International Conference on Learning
  Representations (ICLR)}\/}. San Diego, USA.

\bibitem[{Bastings et~al.(2017)Bastings, Titov, Aziz, Marcheggiani, and
  Simaan}]{BastingsEtAl2017GCN}
Jasmijn Bastings, Ivan Titov, Wilker Aziz, Diego Marcheggiani, and Khalil
  Simaan. 2017.
\newblock Graph convolutional encoders for syntax-aware neural machine
  translation.
\newblock In {\em Proceedings of the 2017 Conference on Empirical Methods in
  Natural Language Processing\/}. Association for Computational Linguistics,
  Copenhagen, Denmark, pages 1947--1957.

\bibitem[{Choi et~al.(2018)Choi, Yoo, and Lee}]{choi2017unsupervisedtree}
Jihun Choi, Kang~Min Yoo, and Sang{-}goo Lee. 2018.
\newblock Unsupervised learning of task-specific tree structures with
  tree-lstms.
\newblock {\em AAAI\/} abs/1707.02786.

\bibitem[{Dozat and Manning(2017)}]{dozat2017biaffine}
Timothy Dozat and Christopher~D. Manning. 2017.
\newblock Deep biaffine attention for neural dependency parsing.
\newblock In {\em {Proceedings of the International Conference on Learning
  Representations (ICLR)}\/}. Toulon, France.

\bibitem[{Eriguchi et~al.(2016)Eriguchi, Hashimoto, and
  Tsuruoka}]{eriguchi2016treetoseq}
Akiko Eriguchi, Kazuma Hashimoto, and Yoshimasa Tsuruoka. 2016.
\newblock Tree-to-sequence attentional neural machine translation.
\newblock In {\em Proceedings of the 54th Annual Meeting of the Association for
  Computational Linguistics (Volume 1: Long Papers)\/}. Association for
  Computational Linguistics, Berlin, Germany, pages 823--833.

\bibitem[{Gehring et~al.(2017)Gehring, Auli, Grangier, and
  Dauphin}]{gehring-EtAl:2017:Long}
Jonas Gehring, Michael Auli, David Grangier, and Yann Dauphin. 2017.
\newblock A convolutional encoder model for neural machine translation.
\newblock In {\em Proceedings of the 55th Annual Meeting of the Association for
  Computational Linguistics (Volume 1: Long Papers)\/}. Association for
  Computational Linguistics, Vancouver, Canada, pages 123--135.

\bibitem[{Hashimoto and Tsuruoka(2017)}]{hashimoto-tsuruoka:2017:EMNLP2017}
Kazuma Hashimoto and Yoshimasa Tsuruoka. 2017.
\newblock Neural machine translation with source-side latent graph parsing.
\newblock In {\em Proceedings of the 2017 Conference on Empirical Methods in
  Natural Language Processing\/}. Association for Computational Linguistics,
  Copenhagen, Denmark, pages 125--135.

\bibitem[{Hochreiter and Schmidhuber(1997)}]{hochreiter97}
Sepp Hochreiter and J{\"u}rgen Schmidhuber. 1997.
\newblock {Long Short-Term Memory}.
\newblock {\em Neural Computation\/} 9(8):1735--1780.

\bibitem[{Jang et~al.(2017)Jang, Gu, and Poole}]{JangEtAl2017:GumbelSoftmax}
Eric Jang, Shixiang Gu, and Ben Poole. 2017.
\newblock Categorical reparameterization with gumbel-softmax.
\newblock In {\em Proceedings of the International Conference on Learning
  Representations (ICLR)\/}. Toulon, France.

\bibitem[{Kim et~al.(2017)Kim, Denton, Hoang, and
  Rush}]{kim2017structuredattention}
Yoon Kim, Carl Denton, Luong Hoang, and Alexander~M. Rush. 2017.
\newblock Structured attention networks.
\newblock In {\em Proceedings of the International Conference on Learning
  Representations (ICLR)\/}. Toulon, France.

\bibitem[{Kingma and Ba(2015)}]{kingma2015adam}
Diederik~P. Kingma and Jimmy Ba. 2015.
\newblock Adam: {A} method for stochastic optimization.
\newblock In {\em {Proceedings of the International Conference on Learning
  Representations (ICLR)}\/}. San Diego, USA.

\bibitem[{Kingma and Welling(2014)}]{Kingma+2014:VAE}
Diederik~P. Kingma and Max Welling. 2014.
\newblock Auto-encoding variational bayes.
\newblock In {\em International Conference on Learning Representations
  (ICLR)\/}. Banff, Canada.

\bibitem[{Luong et~al.(2015)Luong, Pham, and
  Manning}]{luong-pham-manning:2015:EMNLP}
Thang Luong, Hieu Pham, and Christopher~D. Manning. 2015.
\newblock Effective approaches to attention-based neural machine translation.
\newblock In {\em Proceedings of the 2015 Conference on Empirical Methods in
  Natural Language Processing\/}. Association for Computational Linguistics,
  Lisbon, Portugal, pages 1412--1421.

\bibitem[{Maddison et~al.(2017)Maddison, Mnih, and
  Teh}]{MaddisonEtAl2017:Concrete}
Chris~J. Maddison, Andriy Mnih, and Yee~Whye Teh. 2017.
\newblock The concrete distribution: A continous relaxation of discrete random
  variables.
\newblock In {\em {Proceedings of the International Conference on Learning
  Representations (ICLR)}\/}. Toulon, France.

\bibitem[{Maillard et~al.(2017)Maillard, Clark, and
  Yogatama}]{maillard2017jointlylearning}
Jean Maillard, Stephen Clark, and Dani Yogatama. 2017.
\newblock Jointly learning sentence embeddings and syntax with unsupervised
  tree-lstms.
\newblock {\em CoRR\/} abs/1705.09189.

\bibitem[{Marcheggiani and Titov(2017)}]{marcheggiani-titov:2017:EMNLP2017}
Diego Marcheggiani and Ivan Titov. 2017.
\newblock Encoding sentences with graph convolutional networks for semantic
  role labeling.
\newblock In {\em Proceedings of the 2017 Conference on Empirical Methods in
  Natural Language Processing\/}. Association for Computational Linguistics,
  Copenhagen, Denmark, pages 1507--1516.

\bibitem[{Nakazawa et~al.(2016)Nakazawa, Yaguchi, Uchimoto, Utiyama, Sumita,
  Kurohashi, and Isahara}]{nakazawa2016aspec}
Toshiaki Nakazawa, Manabu Yaguchi, Kiyotaka Uchimoto, Masao Utiyama, Eiichiro
  Sumita, Sadao Kurohashi, and Hitoshi Isahara. 2016.
\newblock Aspec: Asian scientific paper excerpt corpus.
\newblock In {\em Proceedings of the Ninth International Conference on Language
  Resources and Evaluation (LREC 2016)\/}. European Language Resources
  Association (ELRA), Portoro{\v z}, Slovenia, pages 2204--2208.

\bibitem[{Ranzato et~al.(2016)Ranzato, Chopra, Auli, and
  Zaremba}]{ranzato2016sequence}
Marc'Aurelio Ranzato, Sumit Chopra, Michael Auli, and Wojciech Zaremba. 2016.
\newblock Sequence level training with recurrent neural networks.
\newblock In {\em {Proceedings of the International Conference on Learning
  Representations (ICLR)}\/}.

\bibitem[{Schuster and Paliwal(1997)}]{schusterpaliwal1997}
Mike Schuster and Kuldip~K. Paliwal. 1997.
\newblock {Bidirectional recurrent neural networks}.
\newblock {\em IEEE Transactions on Signal Processing\/} 45(11):2673--2681.

\bibitem[{Tran and Bisk(2018)}]{tran2018inducing}
Ke~Tran and Yonatan Bisk. 2018.
\newblock \href{https://openreview.net/forum?id=Bkl1uWb0Z}{Inducing grammars
  with and for neural machine translation}.
\newblock
  \href{https://openreview.net/forum?id=Bkl1uWb0Z}{https://openreview.net/forum?id=Bkl1uWb0Z}.

\bibitem[{Vaswani et~al.(2017)Vaswani, Shazeer, Parmar, Jones, Uszkoreit,
  Gomez, and Kaiser}]{vaswani2017attention}
Ashish Vaswani, Noam Shazeer, Niki Parmar, Llion Jones, Jakob Uszkoreit,
  Aidan~N Gomez, and \L~ukasz Kaiser. 2017.
\newblock Attention is all you need.
\newblock In I.~Guyon, U.~V. Luxburg, S.~Bengio, H.~Wallach, R.~Fergus,
  S.~Vishwanathan, and R.~Garnett, editors, {\em Advances in Neural Information
  Processing Systems 30\/}, Curran Associates, Inc., pages 5994--6004.

\bibitem[{Williams et~al.(2017)Williams, Drozdov, and
  Bowman}]{williamsbowman2017isitsyntax}
Adina Williams, Andrew Drozdov, and Samuel~R. Bowman. 2017.
\newblock Learning to parse from a semantic objective: It works. is it syntax?
\newblock {\em CoRR\/} abs/1709.01121.

\bibitem[{Yogatama et~al.(2017)Yogatama, Blunsom, Dyer, Grefenstette, and
  Ling}]{yogatama2017iclr}
Dani Yogatama, Phil Blunsom, Chris Dyer, Edward Grefenstette, and Wang Ling.
  2017.
\newblock Learning to compose words into sentences with reinforcement learning.
\newblock In {\em International Conference on Learning Representations
  (ICLR)\/}. Toulon, France.

\end{thebibliography}
